\documentclass[runningheads]{llncs}
\usepackage[T1]{fontenc}
\usepackage{graphicx}
\usepackage{booktabs}
\usepackage{multirow}
\usepackage{xcolor}
\usepackage{url}

\begin{document}
\title{Do MLLMs Really Understand Low-Resource Khmer Documents? A Pilot Study on Khmer Document VQA}
\titlerunning{Do MLLMs Really Understand Low-Resource Khmer Documents?}
\authorrunning{Thuon N. et al.}
%
\author{Nimol Thuon\inst{1,2}\thanks{Corresponding author}\and
Panhapin Theang\inst{2}}
\institute{University of Science and Technology of China, 230052 Hefei, China \and Université Paris Cité, 75013 Paris, France \\}

%
%
%
\maketitle              
\begin{abstract} 

Recent multimodal large language models (MLLMs) have advanced document understanding, visual question answering, and text extraction. However, their reliability in low-resource, non-Latin settings remains uncertain. Khmer form documents present particular challenges because they contain complex script forms, mixed Khmer--English fields, and monetary values in both Cambodian Riel and US Dollars. Available resources for Khmer Document VQA are also limited. This paper presents a pilot diagnostic evaluation of open MLLMs on Khmer document images. We construct an evaluation subset from the previously introduced KH-FUNSD collection, covering invoices, receipts, quotations, and other business forms. The subset includes questions in English and Khmer, with answers retained in their original English, Khmer, mixed-script, or numeric forms. Rather than introducing a full public benchmark, this study examines the capabilities and failure modes of existing models. We evaluate representative open Qwen-VL models using direct image-based prompting and compare parser-assisted and external OCR-assisted configurations with Qwen3-VL-8B. Direct Qwen3-VL-8B outperforms smaller models, achieving 51.9\% overall accuracy, although performance remains limited for Khmer-script and mixed-script answers. External OCR produces the strongest results, reaching 61.9\% with Tesseract and 61.6\% with PaddleOCR. Nevertheless, Khmer-script answers remain substantially more difficult than English and numeric fields. The results indicate that current MLLMs can process visually clear English and structured numeric content, but reliable native Khmer document understanding remains an open challenge.
\end{abstract}

\section{Introduction}

Multimodal large language models (MLLMs) are increasingly used for document understanding tasks such as visual question answering, OCR-like transcription, information extraction, form interpretation, and layout-aware reasoning. Document Visual Question Answering (Document VQA or DocVQA) has become an important benchmark direction for document intelligence~\cite{mathew2021docvqa,antol2015vqa,souibgui2025docvxqa,mishra2019ocrvqa}.

\begin{figure}[t]
    \centering
    \includegraphics[width=\linewidth]{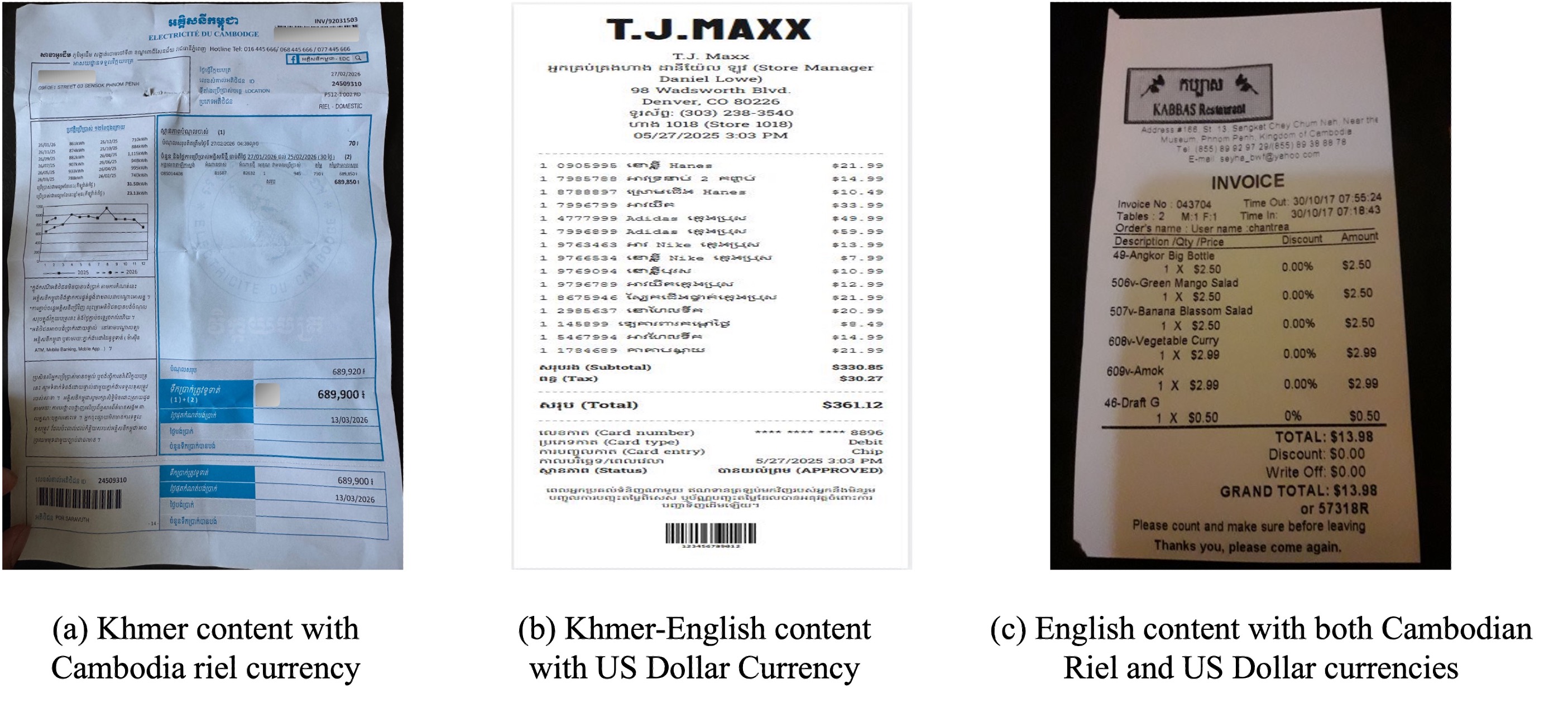}
    \caption{Real-world Khmer business documents present mixed-language and currency challenges: (a) Khmer content with Cambodian Riel currency; (b) mixed Khmer--English content with US Dollar currency; and (c) English-dominant content listing both Cambodian Riel and US Dollar values. These examples illustrate why models must handle native Khmer text, mixed-script fields, and currency grounding simultaneously.}
    \label{fig:khmer_doc_challenges}
\end{figure}

Recent MLLMs, including Qwen-VL~\cite{bai2023qwenvl}, Qwen2.5-VL~\cite{bai2025qwen25vl}, and Qwen3-VL~\cite{bai2025qwen3vl}, suggest that vision-language models are becoming increasingly capable at text-rich image understanding, document-style question answering, and structured information extraction.
\begin{figure}[t]
    \centering
    \includegraphics[width=\linewidth]{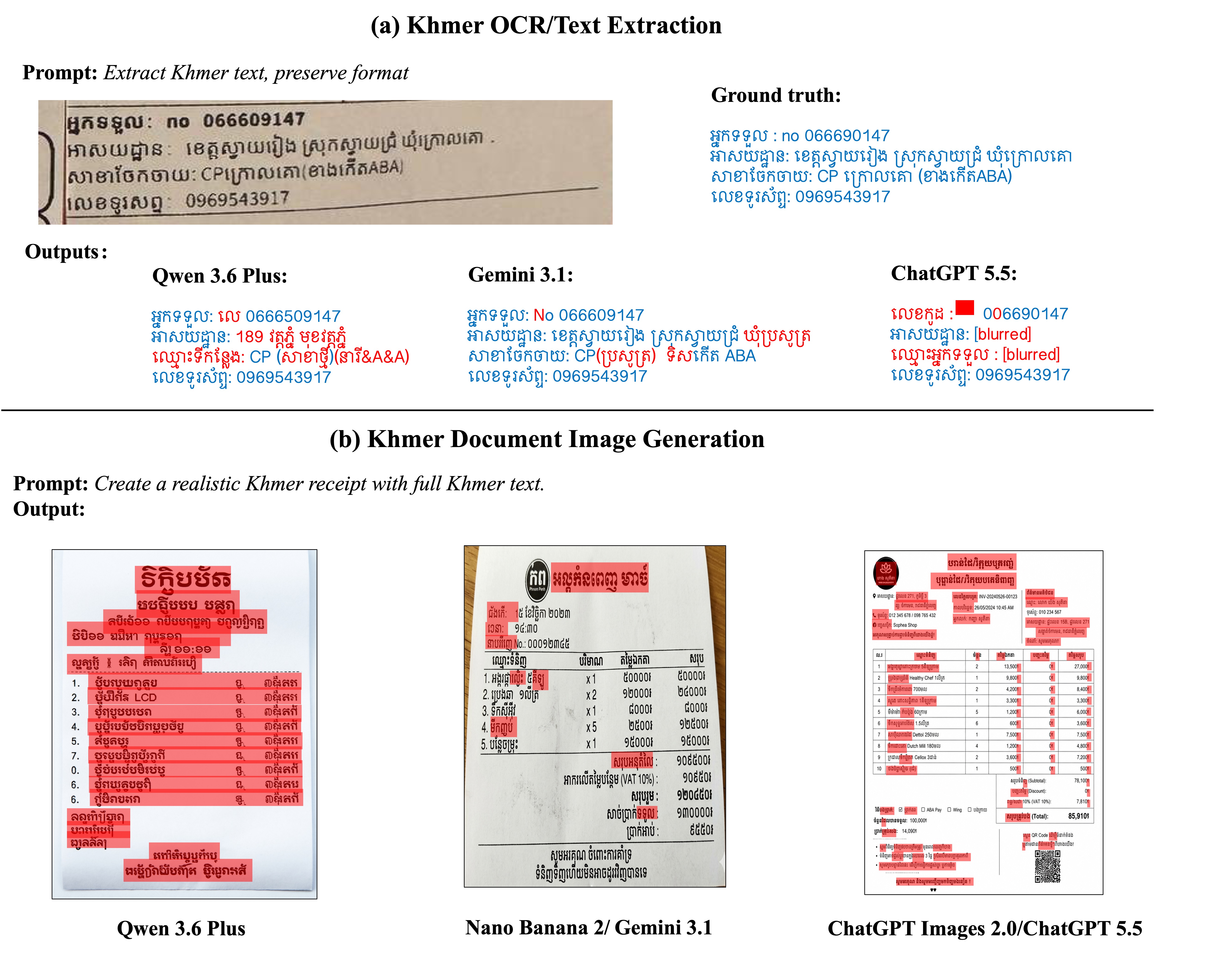}
    \caption{Limitations of current MLLMs on Khmer document content, shown in two tasks: (a) Khmer OCR/text extraction, where models omit or corrupt native Khmer characters; and (b) Khmer document image generation, where models produce visually plausible layouts with pseudo-Khmer or linguistically invalid text.}
    \label{fig:mllm_khmer_failure}
\end{figure}
However, strong progress on mainstream document benchmarks does not necessarily imply robust understanding of low-resource non-Latin documents. Khmer is a low-resource non-Latin language with limited document AI resources compared with high-resource languages such as English, Chinese, Japanese, and major European languages. Khmer script has visually complex characteristics, including stacked consonants, dependent vowels, diacritics, dense character compositions, and limited explicit word separation. These properties make text recognition difficult, particularly in real-world documents affected by blur, low resolution, perspective distortion, compression artifacts, uneven illumination, and handwriting or stamps \cite{thuon2022syllable,keo2025visual,nom2025wildkhmerst}.
For readers unfamiliar with Khmer script, the cropped examples in Fig.~\ref{fig:mllm_khmer_failure} and the document examples in Fig.~\ref{fig:khmer_doc_challenges} illustrate the dense native-script text, stacked forms, and mixed Khmer--English fields that make this setting different from Latin-script document VQA.

Interestingly, current MLLMs may produce plausible outputs for Khmer document tasks while still making serious errors in Khmer text extraction or document generation, as shown in Figure~\ref{fig:mllm_khmer_failure}. This observation motivates the present study. In OCR-like extraction, models may omit words, replace characters, misread numbers, or generate malformed Khmer strings. In document generation, models may create visually plausible receipt or form layouts while producing pseudo-Khmer or mixed Southeast Asian-looking text that is not linguistically valid. These examples indicate that visual plausibility does not necessarily imply genuine Khmer document understanding.
Real-world Khmer documents introduce challenges beyond native script recognition. As shown in Fig.~\ref{fig:khmer_doc_challenges}, Cambodian invoices, receipts, quotations, and administrative forms often contain mixed Khmer--English content. Khmer and English are not always separated into clean monolingual blocks; instead, they may appear in the same line, field, table, product description, company header, or address. Currency usage adds another difficulty. Cambodian business documents may use Cambodian Riel, US Dollars, or both. A model must therefore read the numeric value, identify the associated field, and ground the amount to the correct currency and document context.

Despite these challenges, Khmer document understanding remains underexplored in current Document VQA research. Existing benchmarks have advanced the field, but many are dominated by high-resource languages, relatively clean document images, or document collections that do not fully represent low-resource non-Latin settings. Evaluating MLLMs only on such benchmarks may overestimate their ability to generalize to real-world multilingual documents from underrepresented languages.

In this paper, we present a Khmer-only pilot diagnostic study of MLLM-based Document VQA. Our goal is not to introduce a full benchmark or a broad model leaderboard. Instead, we examine whether recent open MLLMs can reliably answer questions over selected Khmer business document images and analyze where they fail. We design both English-question and Khmer-question settings and preserve answers in their original form. This allows us to analyze model behavior across question language and answer type, including English, Khmer, mixed-script, and number/date/currency answers.

Failure analysis is central to this goal. In business and financial document workflows, a system can appear useful from overall accuracy while still failing on high-risk cases such as names, native Khmer textual fields, or currency amounts. Studying failure cases therefore clarifies what task success would require: not only plausible answers, but reliable native-script reading, field grounding, and calibrated use of OCR evidence.

The main contributions of this paper are as follows:
\begin{itemize}
    \item We present a pilot diagnostic study of MLLM performance on low-resource Khmer Document VQA.
    \item We analyze both English-question and Khmer-question settings while preserving answers in their original script or format.
    \item We compare direct prompting, MLLM-parser-assisted prompting, and external OCR-assisted prompting.
    \item We identify failure modes involving Khmer-script recognition, mixed-script field confusion, currency grounding, and plausible but unsupported answers.
\end{itemize}

\section{Related Work}

\subsection{Document VQA and MLLMs for Document Understanding}

Visual Question Answering (VQA) was introduced as a general vision-language task in which a model answers natural-language questions about images~\cite{antol2015vqa}. Document Visual Question Answering later extended this setting to document images, where models must read text, interpret layout, identify tables or forms, and ground answers in visually structured content. Unlike general image VQA, Document VQA requires close interaction among OCR-like reading, layout understanding, and visual reasoning. The DocVQA dataset introduced this task at scale, with questions defined over document images and answers extracted from document content~\cite{mathew2021docvqa}. Subsequent work extended the setting to document collections, where answering a question may require retrieving relevant documents before extracting the answer~\cite{perez2021doccvqa}. More recent work has also explored explainability in Document VQA, for example by learning visual evidence maps that justify model answers~\cite{souibgui2025docvxqa}.

Recent MLLMs have shown strong progress in image-text understanding and document-style reasoning. Qwen-VL introduced a vision-language model family with text-reading and grounding abilities~\cite{bai2023qwenvl}. Qwen2.5-VL further emphasizes visual recognition, document parsing, structured extraction from forms and tables, and dynamic-resolution processing~\cite{bai2025qwen25vl}. Qwen3-VL extends this family with stronger multimodal reasoning, expanded OCR support, long-context capability, and dense model variants such as 4B and 8B, which are suitable for practical open-model evaluation~\cite{bai2025qwen3vl}.

Despite these advances, document understanding remains vulnerable to fluent but incorrect outputs. Existing Document VQA benchmarks have played an important role in evaluating document understanding systems, but many are still dominated by English or other high-resource language settings. Strong benchmark performance may therefore hide weaknesses on low-resource non-Latin documents. 

\subsection{OCR and Parser-Assisted Document VQA}

Many document understanding pipelines rely on OCR or document parsing before question answering. Early text-centric VQA work showed that text detected by OCR can provide important cues for answering questions over text-rich images~\cite{mishra2019ocrvqa}. In document intelligence, OCR-based representations have also been widely used to combine recognized text, layout positions, and visual features. For example, LayoutLM and LayoutLMv2 model text, layout, and image information jointly for visually rich document understanding tasks~\cite{xu2020layoutlm,xu2021layoutlmv2}. Similarly, DocVQA formulates document visual question answering as answering natural-language questions from document images, often requiring the model to read text and ground answers in document regions~\cite{mathew2021docvqa}.

Recent MLLMs can process document images directly, reducing dependence on explicit OCR pipelines. However, it remains unclear whether direct MLLM prompting is sufficient for low-resource non-Latin scripts. Parser-assisted prompting provides one way to test this issue. A model first produces textual evidence or structured fields from the document, and a second pass uses this evidence to answer questions. External OCR-assisted prompting provides another setting, where text extracted by a dedicated OCR engine is provided to the MLLM together with the image and question.

\subsection{Khmer Document Understanding}

Low-resource document understanding remains challenging because many languages lack large annotated datasets, robust OCR systems, document layout benchmarks, and pretrained language resources. For non-Latin scripts, the problem is further complicated by script-specific visual structures such as diacritics, stacked forms, character shaping, and irregular spacing. These issues become more severe in real-world documents where text may be small, blurred, distorted, stamped, handwritten, or mixed with other languages~\cite{nom2025wildkhmerst,kong2026,thuon2022syllable,thuon2024}.

Khmer document understanding is particularly underrepresented. Prior Khmer OCR studies have highlighted challenges such as stacked characters, diacritics, non-uniform character widths, limited word separation, and scarce training data~\cite{thuon2025kh,nom2025wildkhmerst}. However, most existing work focuses on OCR or scene text recognition rather than Document VQA, where a model must interpret a question, locate the relevant field, read the answer, and preserve the correct script or numeric format.

Real-world Khmer business documents also contain mixed-language and currency rich content. In Cambodia, Khmer is the official language, while English is commonly used in business settings, and other languages such as Chinese may appear in commercial contexts~\cite{tradegov2025cambodia,thuon2025kh}. More generally, multilingual societies often involve code-switching or mixed-language usage~\cite{poplack1980sometimes,auer1998code}. As a result, Cambodian invoices, receipts, and quotations may contain Khmer, English, or mixed Khmer--English fields. They may also include Cambodian Riel, US Dollars, or both, reflecting Cambodia's highly dollarized economy~\cite{odajima2019dollarization,odajima2017dollarization,odajima2019currencychoice}.

\section{Pilot Khmer Document VQA Dataset}

\begin{table}[t]
\centering
\caption{Statistics of the pilot Khmer Document VQA evaluation subset.}
\label{tab:pilot_dataset_statistics}
\scriptsize
\setlength{\tabcolsep}{3pt}
\renewcommand{\arraystretch}{1.1}
\begin{tabular}{@{}p{0.22\columnwidth}p{0.72\columnwidth}@{}}
\toprule
\textbf{Dimension} & \textbf{Categories and QA pairs} \\
\midrule
Document type &
Invoice (168), Quotation (103), Receipt (99) \\

Question language &
English (249), Khmer (121) \\

Answer group &
English (160), Khmer (50), Mixed-script (50), Number/date/currency (110) \\

Question type &
Counting (73), Field value (76), Layout field (78), Name (70), Total amount (73) \\

Difficulty &
Easy (192), Medium (96), Hard (82) \\
\midrule
Total &
62 document images and 370 QA pairs \\
\bottomrule
\end{tabular}
\end{table}

\subsection{Source Collection and Scope}

The pilot evaluation subset is drawn from KH-FUNSD~\cite{thuon2025kh}, a previously introduced collection of real-world Khmer business and administrative documents, including invoices, receipts, quotations, and other forms. These documents contain mixed Khmer--English text, structured fields, tabular layouts, company information, itemized rows, numeric identifiers, dates, and currency values.

We use KH-FUNSD only as the source of a small diagnostic Document VQA subset and do not claim to introduce a full public Khmer Document VQA benchmark. Our goal is to evaluate whether current MLLMs can reliably answer questions about low-resource Khmer documents and to identify common failure patterns.

The subset focuses on invoices, receipts, and quotations because they contain recurring business concepts, structured fields, mixed-language content, and monetary values. They also represent practical Cambodian workflows in which document understanding can support information extraction, search, and administrative processing.

\subsection{Question and Answer Design}

For each document image, we construct question-answer pairs covering common document understanding needs, including field extraction, name lookup, layout-dependent field lookup, counting, and amount-related queries. The evaluation includes English questions because English is commonly used in existing Document VQA benchmarks and allows controlled question writing across documents. To evaluate native-language question understanding, we additionally include Khmer questions translated or adapted from selected English questions while preserving the same answer target.

Answers are preserved in their original form as they appear in the document. Therefore, an answer may be written in English, Khmer, mixed-script form, or number/date/currency format. We keep both raw and normalized answer fields. The raw answer preserves the original document text, while the normalized answer is used only when superficial formatting differences should not affect evaluation, such as commas in numbers, date variants, or currency aliases. We avoid aggressive normalization for Khmer text to prevent distorting native-script answers.

We first describe the composition of the pilot evaluation subset used in this study. Table~\ref{tab:pilot_dataset_statistics} summarizes the pilot subset. It contains 62 document images and 370 question-answer pairs across invoices, receipts, and quotations. The subset is modest in scale, but it is intentionally designed for controlled diagnostic analysis across question language, answer group, question type, and difficulty.
\subsection{Diagnostic Labels}

Each question is assigned a question type and a difficulty label. The question types include counting, field-value extraction, layout-field lookup, name-related questions, and total-amount questions. These categories reflect common operations in business document understanding rather than a complete taxonomy of all possible Document VQA tasks. Difficulty labels are used as diagnostic categories rather than strict cognitive difficulty levels. In this subset, hard questions often require Khmer-script name recognition or native textual-field extraction, so they mainly test native-script reading and grounding rather than complex reasoning alone.

Each question-answer instance contains a document identifier, question identifier, document type, question language, question type, difficulty label, input question, raw answer, normalized answer, answer language, and answer group. These metadata fields enable the analysis reported in Sec.~\ref{sec:results}.

\section{Experimental Setup}
\label{sec:experimental_setup}

\subsection{Question Answering Protocol}
We formulate Khmer Document VQA as a short-answer visual question answering task. Given a document image $I$ and a natural-language question $q$, the model generates a concise answer $a$ based on the visible document content. The question may be written in English or Khmer, and the answer may appear in English, Khmer, mixed-script form, or number/date/currency format.

For all experiments, the model is instructed to return a short JSON object containing the predicted answer and a confidence label:
\begin{verbatim}
{"answer":"...","confidence":"high/medium/low"}
\end{verbatim}
The model is instructed to preserve Khmer, English, numbers, dates, and currency values exactly as they appear whenever possible. If the answer is not visible or uncertain, the model is instructed to return \texttt{Unknown}.

\subsection{Evaluated Models and Evidence Settings}

We evaluate representative open Qwen-VL models under direct image-based prompting. The direct-prompting models are Qwen2.5-VL-3B, Qwen3-VL-4B, and Qwen3-VL-8B. The purpose is not to build a broad leaderboard, but to diagnose whether recent practical MLLMs can reliably answer questions over low-resource Khmer document images.

We further evaluate three Qwen3-VL-8B evidence settings. In the \textbf{direct} setting, the model receives only the document image and the question. In the \textbf{MLLM-parser} setting, Qwen3-VL-8B first generates visible text, key fields, amount/date/currency fields, and layout notes; a second pass then answers using the image, question, and generated parser text. In the \textbf{external OCR} setting, text extracted by Tesseract or PaddleOCR is provided together with the image and question. In all assisted settings, the image remains available to the model and extracted text is treated as supporting evidence rather than a replacement for visual grounding.

\subsection{Implementation and Prompt Details}

All settings use the same JSON output format shown above and ask the model to preserve Khmer, English, numbers, dates, and currency values exactly when possible. In the direct setting, the model receives only the document image and question. In the MLLM-parser setting, Qwen3-VL-8B first generates visible text, key fields, amount/date/currency fields, and layout notes; this generated parser text is then provided to a second answering pass together with the image and question. In the external OCR setting, uncorrected text from Tesseract or PaddleOCR is provided together with the image and question. In all assisted settings, the image remains available, extracted text is treated only as supporting evidence, and no manually corrected or oracle transcript is provided.

OCR engines are used without task-specific fine-tuning or manual transcript correction. The experiments retain the original document images and do not use hand-cropped evidence regions during quantitative evaluation. Generation is run deterministically with sampling disabled. These details are intended to make clear that the assisted settings test whether ordinary parser or OCR evidence improves MLLM answering, rather than whether a fully optimized Khmer OCR pipeline can solve the task.

\subsection{Evaluation Metric}

We report normalized answer accuracy, where a prediction is correct if it matches either the raw or normalized ground-truth answer after basic normalization: English lowercasing, whitespace cleanup, comma removal in numbers, and simple currency aliases. Khmer answers are preserved as written. Accuracy is computed at the QA-pair level over all annotated examples, with unparseable outputs counted as incorrect. We separately report \textbf{Parse}, the percentage of outputs that follow the expected JSON format. We analyze accuracy by question language, answer group, question type, difficulty, and document type.
\section{Results and Analysis}
\label{sec:results}

\subsection{Direct MLLM Performance}

We first evaluate the MLLMs under a direct-prompting setting, where each model receives the document image and question and is asked to generate the answer in the required JSON format. Table~\ref{tab:direct_prompting_results} summarizes the direct-prompting results. Qwen3-VL-8B achieves the best direct-prompting performance, with 51.9\% overall accuracy and 98.6\% parsing success. This indicates that the larger model follows the requested JSON format more reliably and improves general document answering. However, the gains are uneven across answer groups. All direct models perform much better on English answers and number/date/currency fields than on Khmer-script and mixed-script answers. Most notably, all direct models obtain 0.0\% accuracy on Khmer-script answers, suggesting that model scaling alone does not solve the native Khmer text recognition problem.

\begin{table}[t]
\centering
\caption{Direct-prompting results on the Khmer Document VQA pilot set. Accuracy is reported in percentage.}
\label{tab:direct_prompting_results}
\scriptsize
\begin{tabular}{lrrrrrrrr}
\toprule
\multirow{2}{*}{\textbf{Model}} 
& \multirow{2}{*}{\textbf{Parse}} 
& \multirow{2}{*}{\textbf{All}} 
& \multicolumn{2}{c}{\textbf{Question Lang.}} 
& \multicolumn{4}{c}{\textbf{Answer Group}} \\
\cmidrule(lr){4-5} \cmidrule(lr){6-9}
& & 
& \textbf{En-Q} 
& \textbf{Kh-Q} 
& \textbf{En} 
& \textbf{Kh} 
& \textbf{Mix} 
& \textbf{Num.} \\
\midrule
Qwen2.5-VL-3B & 90.5 & 40.8 & 43.4 & 35.5 & 53.8 & 0.0 & 10.0 & 54.5 \\
Qwen3-VL-4B   & 90.3 & 41.9 & 48.2 & 28.9 & 56.3 & 0.0 & 12.0 & 53.6 \\
Qwen3-VL-8B   & 98.6 & 51.9 & 55.4 & 44.6 & 71.3 & 0.0 & 14.0 & 64.5 \\
\midrule
\textbf{Count} & 370 & 370 & 249 & 121 & 160 & 50 & 50 & 110 \\
\bottomrule
\end{tabular}
\end{table}

We further break down direct-prompting performance by question type and difficulty level. Table~\ref{tab:type_difficulty_direct_models} provides a complementary view by question type and difficulty. Qwen3-VL-8B substantially improves counting accuracy, but name-question accuracy remains low. Since many name questions require reading Khmer-script person or organization names, this result is consistent with the answer-group analysis. Accuracy also drops sharply on hard questions. In this pilot subset, hard questions often require native Khmer text recognition rather than complex reasoning alone.

\begin{table}[t]
\centering
\caption{Direct-prompting accuracy by question type and difficulty. Accuracy is reported in percentage over all annotated QA pairs; the Count row gives the total number of annotated QA pairs.}
\label{tab:type_difficulty_direct_models}

\scriptsize
\begin{tabular}{lrrrrrrrr}
\toprule
\multirow{2}{*}{\textbf{Model}}
& \multicolumn{5}{c}{\textbf{Question Type}}
& \multicolumn{3}{c}{\textbf{Difficulty}} \\
\cmidrule(lr){2-6} \cmidrule(lr){7-9}
& \textbf{Count.}
& \textbf{Field value}
& \textbf{Layout}
& \textbf{Name}
& \textbf{Total}
& \textbf{Easy}
& \textbf{Med.}
& \textbf{Hard} \\
\midrule
Qwen2.5-VL-3B & 46.6 & 53.9 & 52.6 & 12.9 & 35.6 & 59.9 & 36.5 & 1.2 \\
Qwen3-VL-4B   & 53.4 & 52.6 & 59.0 & 7.1  & 34.2 & 62.0 & 35.4 & 2.4 \\
Qwen3-VL-8B   & 74.0 & 56.6 & 67.9 & 11.4 & 46.6 & 73.4 & 47.9 & 6.1 \\
\midrule
\textbf{Count} & 73 & 76 & 78 & 70 & 73 & 192 & 96 & 82 \\
\bottomrule
\end{tabular}
\end{table}

\subsection{Parser and OCR-Assisted Evidence}

\begin{figure}[t]
    \centering
    \includegraphics[width=\linewidth]{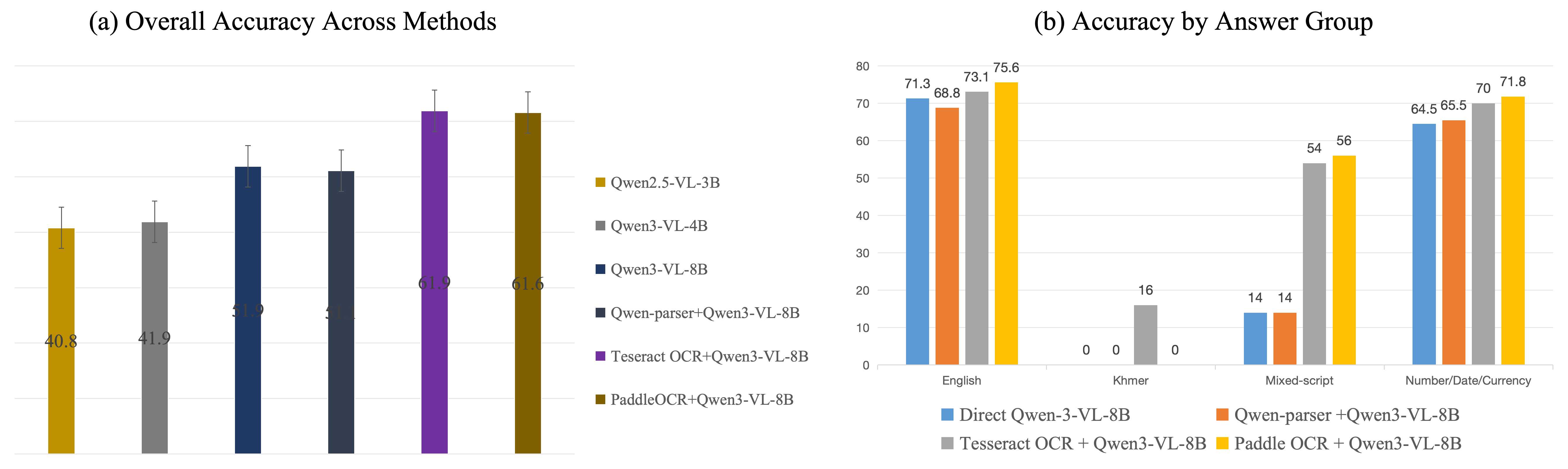}
    \caption{Quantitative results on the pilot Khmer Document VQA subset. (a) Overall accuracy across direct MLLM prompting, MLLM-parser-assisted prompting, and external OCR-assisted prompting, with Wilson 95\% confidence intervals. (b) Accuracy by answer group for Qwen3-VL-8B variants.}
    \label{fig:quantitative_results}
\end{figure}

To assess whether explicit textual evidence helps overcome recognition limitations, we compare direct prompting with parser-assisted and OCR-assisted settings. Figure~\ref{fig:quantitative_results} and Table~\ref{tab:ocr_parser_comparison} summarize these results. Direct Qwen3-VL-8B reaches 51.9\% overall accuracy, while MLLM-parser-assisted prompting remains similar at 51.1\%. This suggests that self-generated parser evidence does not substantially reduce the model's recognition errors. In contrast, Tesseract OCR and PaddleOCR raise overall accuracy to 61.9\% and 61.6\%, respectively.

\begin{table}[t]
\centering
\caption{Direct, parser-assisted, and external OCR-assisted Qwen3-VL-8B results. Accuracy is reported in percentage.}
\label{tab:ocr_parser_comparison}
\scriptsize
\begin{tabular}{lrrrrrrrr}
\toprule
\multirow{2}{*}{\textbf{Method}}
& \multirow{2}{*}{\textbf{Parse}}
& \multirow{2}{*}{\textbf{All}}
& \multicolumn{2}{c}{\textbf{Question Lang.}}
& \multicolumn{4}{c}{\textbf{Answer Group}} \\
\cmidrule(lr){4-5} \cmidrule(lr){6-9}
& & & \textbf{En-Q} & \textbf{Kh-Q}
& \textbf{En} & \textbf{Kh} & \textbf{Mix} & \textbf{Num.} \\
\midrule
Direct & 98.6 & 51.9 & 55.4 & 44.6 & 71.3 & 0.0 & 14.0 & 64.5 \\
MLLM-parser & 99.2 & 51.1 & 53.4 & 46.3 & 68.8 & 0.0 & 14.0 & 65.5 \\
Tess. OCR & \textbf{99.7} & \textbf{61.9} & \textbf{64.3} & 57.0 & 73.1 & \textbf{16.0} & 54.0 & 70.0 \\
PaddleOCR & 98.1 & 61.6 & 63.5 & \textbf{57.9} & \textbf{75.6} & 0.0 & \textbf{56.0} & \textbf{71.8} \\
\bottomrule
\end{tabular}
\end{table}

The improvement is especially large for mixed-script and number/date/currency answers. PaddleOCR improves mixed-script accuracy from 14.0\% to 56.0\%, while Tesseract reaches 54.0\%. However, native Khmer-script answers remain difficult. Tesseract improves Khmer-answer accuracy to 16.0\%, but PaddleOCR remains at 0.0\%. This contrast suggests that external OCR is useful for structured and mixed-script evidence, but low-resource Khmer recognition remains unresolved.

The confidence intervals emphasize that small differences should not be overinterpreted. For example, the overall direct Qwen3-VL-8B result of 51.9\% has a Wilson 95\% interval of approximately 46.8--56.9\%, while Tesseract OCR and PaddleOCR have intervals of approximately 56.8--66.7\% and 56.6--66.4\%, respectively. For 50-example answer groups, uncertainty is much larger: the direct mixed-script result of 14.0\% corresponds to roughly 6.9--26.2\%, while the PaddleOCR mixed-script result of 56.0\% corresponds to roughly 42.3--68.8\%. Thus, the main conclusion is not a fine ranking between OCR systems, but the broader pattern that OCR evidence helps mixed-script and numeric fields while native Khmer answers remain difficult.

\subsection{Question-Type Gains from OCR}

We further analyze the effect of OCR assistance across different question categories. Table~\ref{tab:ocr_question_type} shows that OCR assistance changes which question types can be answered. Tesseract provides the largest improvement for name questions, increasing accuracy from 11.4\% to 37.1\%. PaddleOCR gives the strongest result for total-amount questions, increasing accuracy from 46.6\% to 61.6\%. These gains suggest that OCR evidence helps the model locate and copy difficult textual or numeric fields, although name questions remain substantially harder than counting and field-value questions.

\begin{table}[t]
\centering
\caption{Question-type comparison for Qwen3-VL-8B under direct and OCR-assisted settings. Accuracy is reported in percentage over all annotated QA pairs; the Count row gives the total number of annotated QA pairs.}
\label{tab:ocr_question_type}
\scriptsize

\begin{tabular}{lrrrrr}
\toprule
\multirow{2}{*}{\textbf{Method}}
& \multicolumn{5}{c}{\textbf{Question Type}} \\
\cmidrule(lr){2-6}
& \textbf{Count.}
& \textbf{Field value}
& \textbf{Layout}
& \textbf{Name}
& \textbf{Total} \\
\midrule
Direct & 74.0 & 56.6 & 67.9 & 11.4 & 46.6 \\
Tess. OCR & 74.0 & 68.4 & \textbf{73.1} & \textbf{37.1} & 54.8 \\
PaddleOCR & \textbf{80.8} & \textbf{69.7} & 64.1 & 30.0 & \textbf{61.6} \\
\midrule
\textbf{Count} & 73 & 76 & 78 & 70 & 73 \\
\bottomrule
\end{tabular}
\end{table}

\subsection{Qualitative Failure Analysis}

\begin{figure}[t]
    \centering
    \includegraphics[width=\linewidth]{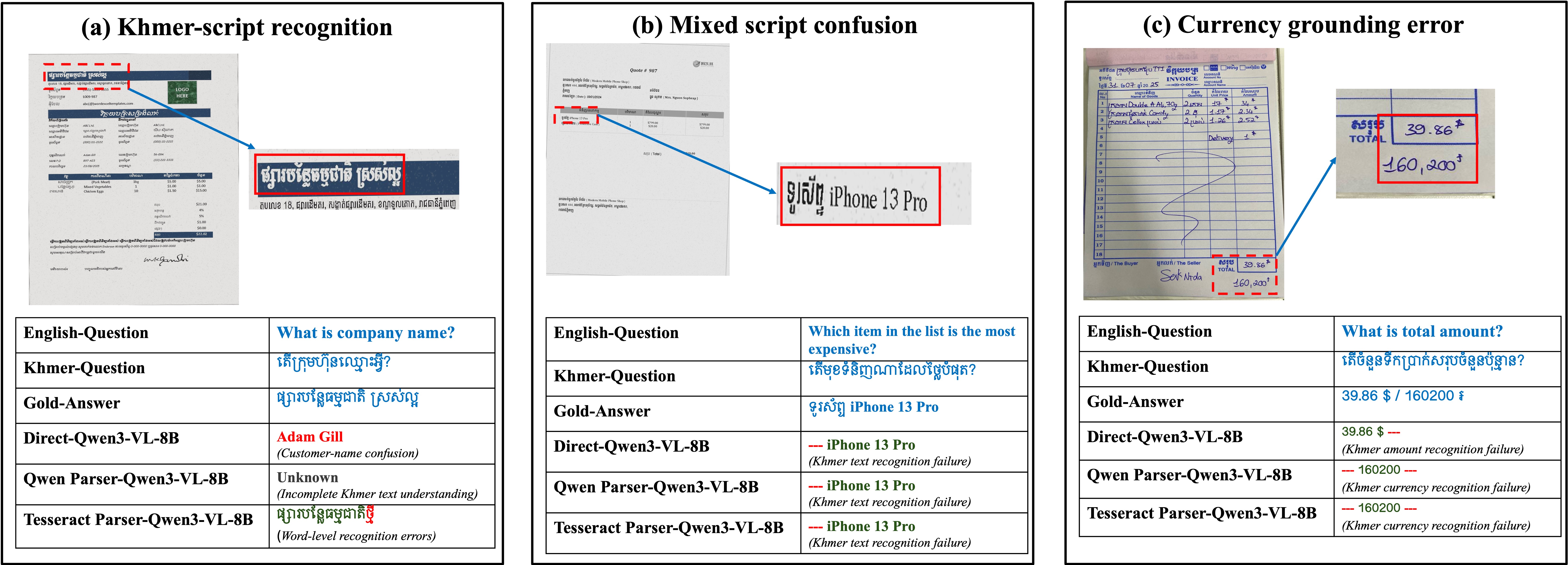}
    \caption{Representative failure modes in Khmer Document VQA. (a) Khmer-script recognition errors occur when the model fails to read native Khmer names or textual fields. (b) Mixed-script confusion occurs when Khmer and English appear in the same field, causing the model to copy only the Latin-script portion or produce a partial answer. (c) Currency grounding errors occur when the model reads nearby numeric values but associates them with the wrong field or currency. Green text indicates correct answers, red text indicates incorrect predictions, and cropped regions highlight the relevant evidence.}
    \label{fig:qualitative_failures}
\end{figure}

\begin{figure}[t]
    \centering
    \includegraphics[width=\linewidth]{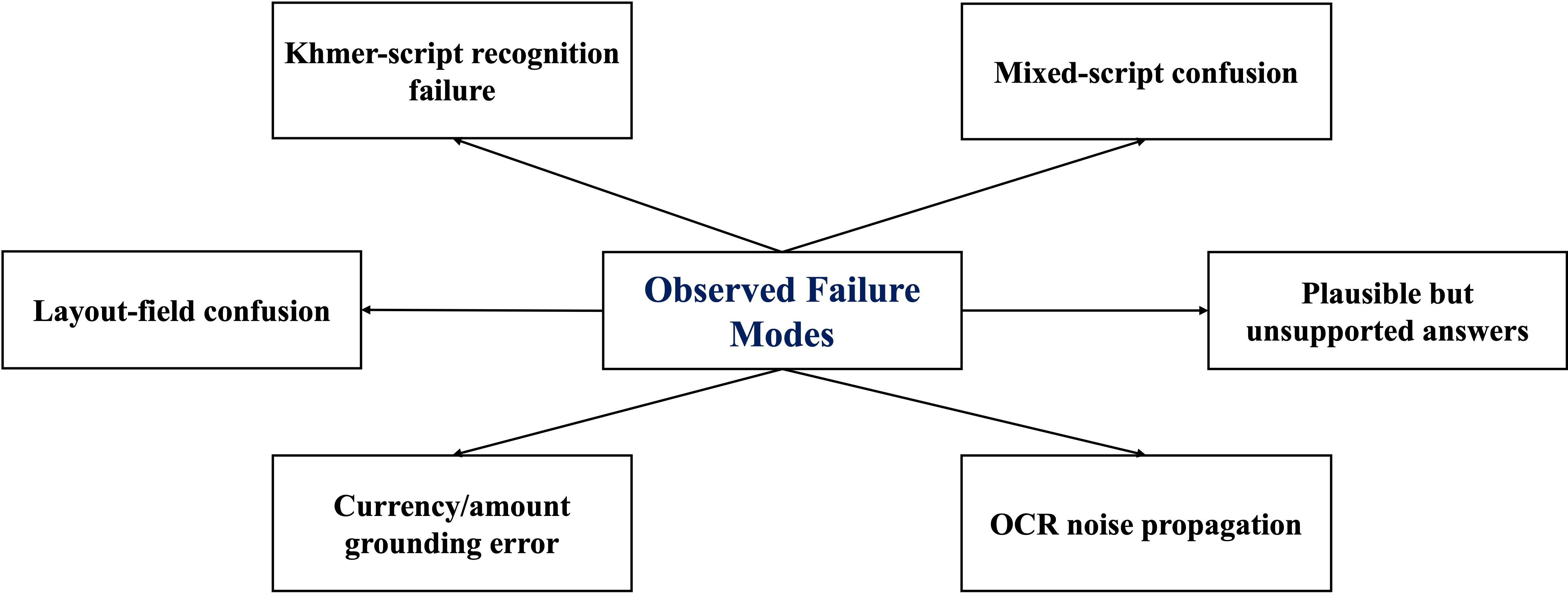}
    \caption{Observed failure modes in Khmer Document VQA, including Khmer-script recognition failure, mixed-script confusion, layout-field confusion, currency/amount grounding errors, OCR noise propagation, and plausible but unsupported answers.}
    \label{fig:failure_taxonomy}
\end{figure}

To better interpret the quantitative results, we examine representative failure cases produced by the MLLMs. As shown in Figure~\ref{fig:failure_taxonomy}, the observed failures can be grouped into several major categories, including Khmer-script recognition errors, mixed-script confusion, layout-field confusion, currency or amount grounding errors, OCR noise propagation, and plausible but unsupported answers. Additionally, Figure~\ref{fig:qualitative_failures} provides qualitative examples that further explain these error patterns. The first example shows a Khmer-script recognition failure, where the model cannot correctly read a native Khmer textual field. The second example shows mixed-script confusion: the English part of the field is easier to copy, while the Khmer component is ignored or corrupted. The third example shows a currency grounding error, where the model reads a nearby amount but associates it with the wrong currency or total field. These examples support the main finding that current MLLMs are limited not only by answer formatting, but also by native-script recognition and document-field grounding.

\section{Discussion and Limitations}
\label{sec:discussion_limitations}

\subsection{Discussion}

The results support four main findings. Larger Qwen-VL models improve direct accuracy and output-format stability, but direct prompting remains unreliable for Khmer-script and mixed-script answers. The bottleneck is not only output formatting: Qwen3-VL-8B parses reliably but still misses many native-script fields. MLLM-generated parser evidence gives little gain, suggesting that self-parsing often repeats the same recognition errors. External OCR provides the strongest improvement, especially for mixed-script and number/date/currency fields, yet Khmer-script answers remain much harder than English and numeric fields.

Overall, current MLLMs can handle some English-visible and structured numeric content in Khmer documents, especially with OCR support. However, robust native Khmer document understanding remains unresolved, and aggregate accuracy can hide severe native-script failures. Script-aware and language-aware evaluation is therefore necessary for low-resource Document VQA.

\subsection{Limitations and Future Work}

This study is a pilot diagnostic study rather than a full benchmark. It uses 62 document images and 370 question-answer pairs from one Khmer form-understanding source collection, so coverage across document domains is limited. The experiments also focus on open Qwen-VL models and two OCR engines. Future work should evaluate broader model families, larger MLLMs, closed-source systems such as GPT-4o or Gemini, and additional OCR systems.

The evaluation protocol also has limits. Normalized exact matching is practical for short-answer Document VQA but can mis-handle equivalent Khmer text, names, or currency formats. We do not include an oracle OCR condition with human transcripts, so OCR recognition errors cannot be fully separated from downstream MLLM comprehension or grounding errors. Confidence labels are collected but not yet calibrated. Future work should add human adjudication, oracle-transcript experiments, confidence calibration, Khmer-aware normalization, and a larger public Khmer or multilingual non-Latin Document VQA benchmark, subject to privacy and copyright constraints.

\section{Conclusion}

This paper presented a pilot diagnostic study of current open MLLMs for Khmer Document VQA, an underexplored low-resource non-Latin document setting. Using 62 Khmer document images and 370 question-answer pairs, we evaluated direct Qwen-VL prompting, MLLM-parser-assisted prompting, and external OCR-assisted prompting across question language, answer group, question type, difficulty, and evidence setting. The results show that Qwen3-VL-8B improves direct performance and output-format stability over smaller models, but remains weak on Khmer-script and mixed-script answers. MLLM-generated parser evidence provides little improvement, suggesting that self-parsing often repeats the same recognition errors. In contrast, external OCR assistance improves performance, especially for mixed-script and number/date/currency fields, although native Khmer-script answers remain difficult. Overall, current MLLMs are not yet robust Khmer document readers. They can handle some structured, English-visible, and numeric content, but native-script recognition, mixed-script grounding, and currency-field interpretation remain major challenges. Future work should develop larger Khmer and multilingual non-Latin Document VQA benchmarks, stronger Khmer-aware OCR and normalization methods, and evaluation protocols that explicitly measure native-script and mixed-language document understanding.

\bibliographystyle{splncs04}
\bibliography{references}
\end{document}